\documentclass[10pt,twocolumn,letterpaper]{article}

\usepackage{cvpr}
\usepackage{times}
\usepackage{epsfig}
\usepackage{graphicx}
\usepackage{amsmath}
\usepackage{amssymb}

\usepackage{multicol}
\usepackage{authblk}
\usepackage{verbatim}
\usepackage{caption}
\usepackage{algorithm,algcompatible,amsmath}
\usepackage{booktabs} 
\algnewcommand\INPUT{\item[\textbf{Input:}]}%
\algnewcommand\OUTPUT{\item[\textbf{Output:}]}%
\usepackage{pbox}
\usepackage{adjustbox}
\usepackage{colortbl}
\definecolor{Gray}{gray}{0.85}
\newcolumntype{a}{>{\columncolor{Gray}}c}
\newcolumntype{b}{>{\columncolor{white}}c}
\usepackage{hyperref}
\cvprfinalcopy

\begin{document}

%%%%%%%%% TITLE
\title{Adversarial Semantic Hallucination for Domain Generalized Semantic Segmentation}
\author[1]{Gabriel Tjio}
\author[1, *]{Ping Liu }
\author[1]{Joey Tianyi Zhou}
\author[1]{Rick Siow Mong Goh}
\affil[1]{ Institute of High Performance Computing, A*STAR, Singapore}
\affil[ ]{\textit{gabriel-tjio@ihpc.a-star.edu.sg; pino.pingliu@gmail.com; \{joey\_zhou;gohsm\}@ihpc.a-star.edu.sg}}
\affil[*]{corresponding author}
\maketitle
%\thispagestyle{empty}
%%%%%%%%% ABSTRACT
\begin{abstract}
Convolutional neural networks typically perform poorly when the test (target domain) and training (source domain) data have significantly different distributions. While this problem can be mitigated by using the target domain data to align the source and target domain feature representations, the target domain data may be unavailable due to privacy concerns. Consequently, there is a need for methods that generalize well despite restricted access to target domain data during training. In this work, we propose an adversarial semantic hallucination approach (ASH), which combines a class-conditioned hallucination module and a semantic segmentation module. Since the segmentation performance varies across different classes, we design a semantic-conditioned style hallucination module to generate affine transformation parameters from semantic information in the segmentation probability maps of the source domain image.  Unlike previous adaptation approaches, which treat all classes equally, ASH considers the class-wise differences. The segmentation module and the hallucination module compete adversarially, with the hallucination module generating increasingly ``difficult" stylized images to challenge the segmentation module. In response, the segmentation module improves as it is trained with generated samples at an appropriate class-wise difficulty level. Our results on the Cityscapes and Mapillary benchmark datasets show that our method is competitive with state of the art work. Code is made available at \url{https://github.com/gabriel-tjio/ASH}.
\end{abstract}

%%%%%%%%% BODY TEXT
\section{Introduction}

\begin{figure}
\centering
\includegraphics[width=0.47\textwidth,keepaspectratio]{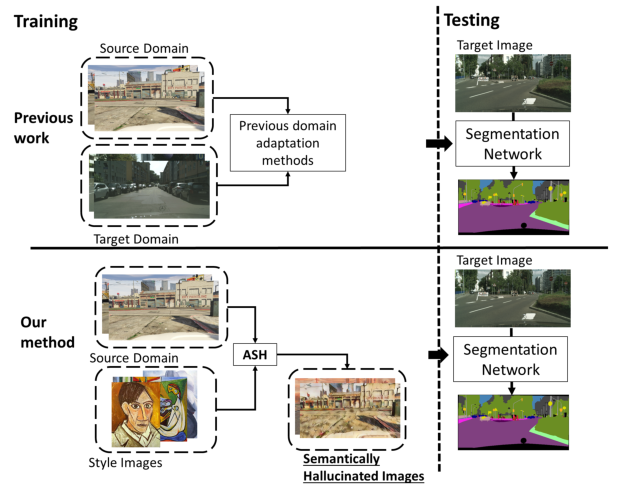}
\caption{Illustrated summary of our proposed Adversarial Semantic Hallucination approach (ASH). Previous domain adaptation works require target domain data during training. Since target domain data are unavailable in our problem setting, we generate additional data with randomized styles via style transfer with ASH.}
\label{figure:ASHintro}

\end{figure}

Semantic segmentation \cite{7913730} involves classifying image pixels into a given category. While deep learning has vastly improved semantic segmentation performance, it requires large amounts of pixel-level annotated data. Pixel-level annotation is time-consuming and error-prone, making it impractical for real-life applications. For training vision systems in autonomous vehicles, synthetic data are readily available and easily labeled. However, synthetic data (source domain data) differ visually from real-world driving data (target domain data). As a result of this domain gap, models that are trained solely on synthetic data perform poorly on real-world data. 

Domain adaptation methods \cite{conf/cvpr/BousmalisSDEK17,DBLP:series/acvpr/GaninUAGLLML17,pmlr-v80-hoffman18a, Luo_2019_ICCV, Saito_2018_CVPR, Luo_2019_CVPR,Yang_eccv2020_label} seek to minimize the domain gap between the source domain and target domain by utilizing unlabeled target domain data. Unfortunately, in Domain Generalization (DG) scenarios \cite{Yue_2019_ICCV,ieee_tip_duo_texture_rand,Choi_2021_CVPR,chen2021contrastive}, target domain data are not accessible during training. With limited access to target data, it becomes quite difficult, if not impossible, to apply previous unsupervised domain adaptation methods \cite{conf/cvpr/BousmalisSDEK17,DBLP:series/acvpr/GaninUAGLLML17,pmlr-v80-hoffman18a, Luo_2019_ICCV, Saito_2018_CVPR, Luo_2019_CVPR,Yang_eccv2020_label}. To solve this problem, hallucination-based approaches \cite{Kim_2020_CVPR,Yue_2019_ICCV,DBLP:conf/nips/LuoLGY020} have been proposed. These methods generate augmented images by varying texture information in the source domain images. By randomizing these domain variant features, the trained model becomes more sensitive to domain invariant features, such as shape information. The increased sensitivity to domain invariant features helps the model generalize better to the unseen target domain data. For example, Adversarial Style Mining \cite{DBLP:conf/nips/LuoLGY020} (ASM) uses a single target domain image to hallucinate additional training images. The global statistics of the single target domain image are used to adaptively stylize the source domain images. The ``difficulty" of the stylized images is progressively increased via adversarial training.

Most prior works \cite{Kim_2020_CVPR,Yue_2019_ICCV,DBLP:conf/nips/LuoLGY020} conduct hallucination in a global manner and fail to consider the statistical differences between different classes. In real scenarios, datasets might be imbalanced because of collection and/or annotation difficulties. Consequently, classes with fewer examples are more difficult to predict accurately. For example, in the driving datasets \cite{Richter_2016_ECCV,RosCVPR16,Cordts2016Cityscapes}, a larger proportion of pixels correspond to ``road", ``building", or ``sky" classes compared to minority classes such as ``pole" or ``light". We argue that uniformly stylizing all classes without considering their different characteristics may lead to a sub-optimal result.

% This can be problematic because datasets might be imbalanced due to the collection and/or annotation difficulties, making some classes more difficult. For example, in the driving datasets \cite{Richter_2016_ECCV,RosCVPR16,Cordts2016Cityscapes}, a larger proportion of pixels correspond to ``road", ``building", or ``sky" classes compared to minority classes such as ``pole" or ``light" for each image. Applying the same degree of stylization to all classes without considering their different characteristics may lead a biased model.

% reduce performance for minority classes. 

Prior works, such as \cite{Lin_2017_ICCV, 8953804}, tried to address this problem by leveraging focal loss \cite{Lin_2017_ICCV} or class balanced loss \cite{8953804}. However, these approaches still have their limitations. Class balancing methods like focal loss \cite{Lin_2017_ICCV} assume that source and target domain distributions are similar, which does not always hold true \cite{Jamal_2020_CVPR}. Additionally, hyperparameter selection for these methods \cite{Lin_2017_ICCV,8953804} is nontrivial and the hyperparameters may not be transferable between datasets. 

To address these limitations, we propose a new method, Adversarial Semantic Hallucination (ASH), for domain generalized semantic segmentation. Inspired by ASM  \cite{DBLP:conf/nips/LuoLGY020}, we further extend it by using semantic information to guide adversarial hallucination and improve generalizability. The semantic information from the segmentation probability map is used to differentiate between classes based on their segmentation difficulty and generate transformation parameters for the style features.  ASH stylizes the source domain images with these transformation parameters. Next, ASH collaborates with a discriminator in an adversarial manner by adaptively generating challenging data for training the segmentation network. With our method, the segmentation network not only becomes better at differentiating between classes, but also demonstrates good generalizability across different domains.

Our main contributions are summarized as follows:

1) We present ASH for domain generalized semantic segmentation. ASH leverages semantic information to conduct a class-conditioned stylization for source domain images, making the trained model generalize better. Unlike previous work such as ASM \cite{DBLP:conf/nips/LuoLGY020} which utilizes stylization, our method does not need any target domain data during training and thus is more practical. Additionally, our approach also considers the different characteristics between classes instead of treating them equally.

2) We conduct extensive domain generalized semantic segmentation experiments to test the efficacy of ASH, including domain generalization from GTA5 \cite{Richter_2016_ECCV} or SYNTHIA datasets \cite{RosCVPR16} $\rightarrow$ the Cityscapes \cite{Cordts2016Cityscapes} or Mapillary benchmark datasets \cite{mapillary_dataset}. The experimental results demonstrate the efficacy of ASH even when target data are not available during training.

\section{Related Work}
In this section, we briefly survey previous works that are most related to ours, including unsupervised domain adaptation and generative adversarial networks.

\subsection{Unsupervised Domain Adaptation}
Unsupervised domain adaptation (UDA) is a subset of transfer learning. Given labeled source data and unlabeled target data, UDA aims to train a network to achieve satisfactory performance on target domain data. Previous works \cite{Saito_2018_CVPR,DBLP:series/acvpr/GaninUAGLLML17} align the feature representations of the source and target domains by minimizing the discrepancy between the two domains. Following this alignment approach, the knowledge learned from the source domain can be applied to the target domain. UDA methods can be generally divided into three categories, namely pixel-level alignment, feature-level alignment, and output-level alignment. Pixel-level domain adaptation \cite{conf/cvpr/BousmalisSDEK17} transforms the source domain images to visually mimic the target domain images. The transformed source domain images are included during training. Alternatively, target-to-source image translation has also been explored \cite{Yang_eccv2020_label}. Different from these approaches, our method reduces overfitting to textural features in the source domain data instead of generating data that mimics either domain. Feature-level domain
adaptation \cite{DBLP:series/acvpr/GaninUAGLLML17,pmlr-v80-hoffman18a, Luo_2019_ICCV} aligns the feature representations across domains, making the feature representations extracted from the source and target domain indistinguishable. This approach has been studied for image classification \cite{DBLP:series/acvpr/GaninUAGLLML17} and semantic segmentation \cite{Luo_2019_ICCV}. Output-level domain adaptation \cite{Saito_2018_CVPR, Luo_2019_CVPR} maximises the similarity between domains at the output level. Tsai \etal\cite{Tsai_adaptseg_2018} and Luo \etal \cite{9372870} demonstrated that output-level alignment performs better compared to feature-level alignment for semantic segmentation. Recently, source-free adaptation methods such as \cite{kundu_eccv2020} adapt a model pretrained on source domain data to the target domain. The problem setting for such work restricts access to source domain data instead of target domain data after pretraining the model. In contrast, our method does not use target domain data during training.
The work most related to our approach is ASM \cite{DBLP:conf/nips/LuoLGY020}. Luo \textit{et al.} \cite{DBLP:conf/nips/LuoLGY020} aim to solve unsupervised domain adaptive semantic segmentation when limited unlabeled target data are available. Both ASM \cite{DBLP:conf/nips/LuoLGY020} and our approach utilize a style transfer strategy to generate augmented data. However, there are significant differences between the two works: (1) ASM requires target data (one single target domain image) for domain alignment. Conversely, our approach does not need any target data for training, making it more applicable for real-life scenarios. (2) ASM uses a global stylization approach. The stylized image is globally updated with the target task prediction loss on the stylized data. Consequently, pixels from different classes are uniformly stylized, which could reduce performance for the `harder' classes on target domain data. In contrast, we consider the class-wise differences and utilize the semantic information for a class-conditioned process. The experimental results reported in Tables.\ref{tab:gta-cityscapes} and \ref{tab:synthia-cityscapes}  demonstrate the advantages of ASH compared to ASM.
\subsection{Generative Adversarial Networks (GANs)} 
GANs have garnered much attention since their introduction \cite{NIPS2014_5ca3e9b1} and have been used in a wide range of applications, such as image generation \cite{Karras_2019_CVPR} and image translation \cite{CycleGAN2017}. GAN architecture typically comprises of a generator-discriminator pair optimized in a min-max fashion. The generator is trained to synthesize realistic images while the discriminator is trained to distinguish between the synthesized images and the real images.
Though GANs have been used for unsupervised domain adaptation \cite{Luo_2019_CVPR,conf/cvpr/BousmalisSDEK17}, the lack of target domain data for the domain generalization problem setting means that some modifications are required. Therefore, we train the discriminator to distinguish between segmented output from the source domain images and the randomly stylized source domain images.

Next, we apply the principle behind conditional GANs \cite{8579015} for greater control over the stylization extent of the source domain image. Conditional GANs give the user additional control over the generated output via prior information to the generator. We were also further inspired by recent works \cite{park2019SPADE,wang2018sftgan} which demonstrate prior information improves synthesized image quality. Wang \etal \cite{wang2018sftgan} leverage semantic information to improve output image quality during super- resolution. The probability map serves as a prior and is used as an input for spatial transformation of the image features. Similarly, Park \etal \cite{park2019SPADE} condition the synthesized GAN output with semantic information during feature transformation. This enables their approach to generate realistic images, while also allowing the user to determine the content of the generated images.  

We extend existing domain adaptation work by incorporating semantic information as a prior. %While Luo \etal\cite{DBLP:conf/nips/LuoLGY020} uses an adversarial based approach to train their network, we include semantic information to regularize the extracted style features.%
Our ASH module is lightweight and only consists of a few convolutional layers to: 1) map the semantic information to latent space, and 2) compute the transformation coefficients for the style features. Furthermore, ASH is required only during training and therefore does not increase computation cost during inference.

% In addition to minimizing the computational costs during training, the ASH module is not required during inference. Therefore, it does not increase the computational cost nor require additional parameters during model evaluation.  

\section{Method}
In this section, we firstly discuss our problem setting and preliminary background. We then provide the technical details for ASH.

\begin{figure*}[t]
\begin{center}
\includegraphics[width=1.0\linewidth,keepaspectratio]{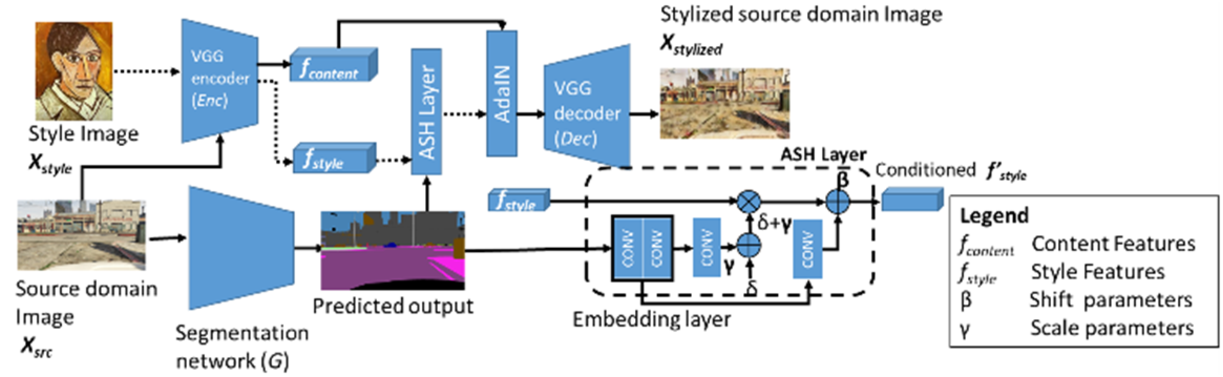}
\end{center}
  \vspace{-0.5cm}
\caption{Illustrated workflow for generating stylized source domain images with Adversarial Semantic Hallucination (ASH). A pretrained VGG encoder extracts features from the source and style images. ASH conditions the style features with semantic information from the segmented source domain image. The semantic information is used to generate the element-wise scale and shift parameters $\boldsymbol{\gamma}$ and $\boldsymbol{\beta}$. These transformation parameters adjust the style features based on the predicted class in the segmentation output. Since the transformation parameters are only intended to perturb the style features, we include a non-zero constant $\delta$.  The content features are re-normalized with the transformed style features via Adaptive Instance Normalization (AdaIN). The merged features are then decoded to output stylized source images.}
\label{figure:workflow_ash}
\end{figure*}

\subsection{Problem Setting}
The problem setting for domain generalization is defined as follows: We have source domain data $\boldsymbol{X_{src}}$ with labels $\boldsymbol{Y_{src}}$ during training, but we cannot access target domain data $\boldsymbol{X_{target}}$. The source domain and target domain have different data distributions (i.e P($\boldsymbol{X_{src}}$,$\boldsymbol{Y_{src}}$) $\neq$ P($\boldsymbol{X_{target}}$,$\boldsymbol{Y_{target}}$)). Our goal is to develop a model $G$ that correctly predicts the target domain labels after training. 

\subsection{Preliminary background}
Our method can be divided into 2 stages. In the first stage, our approach incorporates the style transfer method \cite{huang2017adain}. We augment the source domain data $\boldsymbol{X_{src}}$ by stylizing it with images from a paintings dataset $\boldsymbol{X_{sty}}$, \textit{i.e.}, Painter by Numbers. The style features are conditioned with semantic information obtained from the segmentation output of source domain data. In the second stage, we separately train the different components: an ASH module, a segmentation network and a discriminator.

Similar to \cite{huang2017adain}, we use a pretrained VGG19 network to extract features from the source domain images and style images. We then use adaptive instance normalization \cite{huang2017adain}:

\small \begin{equation}
\text{AdaIN}(\boldsymbol{f_{src}}, \boldsymbol{f_{sty}}) = \sigma(\boldsymbol{f_{sty}}) \left( \frac{\boldsymbol{f_{src}} - \mu(\boldsymbol{f_{src}})}{\sigma(\boldsymbol{f_{src}})} \right) + \mu(\boldsymbol{f_{sty}}) 
\label{eqn:adain}
\end{equation}
\normalsize
which re-normalizes the channel-wise mean $\mu(.)$ and variance $\sigma(.)$ of the content features (i.e source features $\boldsymbol{f_{src}}$) to match that of the style features $\boldsymbol{f_{sty}}$. 

Firstly, we aim to improve the generalizability of the trained segmentation model by introducing randomized texture variations during training. At each iteration, we randomly select a style image to stylize the source image. By stylizing the source image with randomized style information, the model learns to disregard texture information.  

Next, we increase the diversity of the style features by introducing orthogonal noise \cite{wang2020diversified}. Orthogonal noise allows us to preserve the inner product of the style features, or its ``inherent style information",  while simultaneously increasing the diversity of the style features \cite{wang2020diversified}. We regularize the segmentation output with label smoothing before conditioning the style features with the ASH module. 

\subsection{Adversarial Semantic Hallucination}
As shown in Figure \ref{figure:workflow_ash}, our framework comprises a segmentation network, a discriminator and an ASH module. The ASH module conditions the style features with semantic information from the source data segmentation output. 

Prior hallucination works such as \cite{DBLP:conf/nips/LuoLGY020} conduct the stylization without considering class-wise differences, which might be sub-optimal. We take a different approach by conducting the hallucination conditioned on semantic information. The semantic information is used to compute the scale $\pmb{\gamma}$ and shift $\pmb{\beta}$ transformation parameters. These parameters transform the style features in latent space. Depending on the predicted class for each pixel, ASH is trained to maximize adversarial loss by assigning different scale and shift transformation parameters. We use adaptive instance normalization \cite{huang2017adain} to merge the content features with the transformed style features. 

\begin{algorithm*}
  \caption{Adversarial approach for domain generalization}
 \begin{algorithmic}[1]
  \INPUT Source domain data $\boldsymbol{X_{src}}$, Source domain label $\boldsymbol{Y_{src}}$, Style image $\boldsymbol{X_{sty}}$, Segmentation network \textit {G}, Discriminator \textit {D}, Encoder \textit {Enc}, Decoder \textit {Dec}, Adversarial Semantic Hallucination ASH, Adaptive Instance Normalization (AdaIN), Number of iterations $Iter_{num}$
  \OUTPUT Optimized segmentation network for domain generalization
  \FOR{$0$, ..., $Iter_{num}$}
   \STATE Generate source features $\boldsymbol{f_{src}}$  with pretrained encoder $Enc(\boldsymbol{X_{src}})$.
   \STATE Generate style features $\boldsymbol{f_{style}}$ with pretrained encoder $Enc(\boldsymbol{X_{sty}})$. 
   \STATE Multiply style features $\boldsymbol{f_{sty}}$ with an orthogonal noise matrix $\boldsymbol{Z}$ sampled from a normal distribution 
   \STATE Obtain scale $\pmb{\gamma}$ and shift $\pmb{\beta}$ coefficients from segmentation output $G(\boldsymbol{X_{src}})$
   \STATE Generate perturbed style features $\boldsymbol{f_{sty}^{'}}$ from $\boldsymbol{f_{sty}}$ with scale $\pmb{\gamma}$ and shift $\pmb{\beta}$ coefficients.   
   \STATE Derive merged source features $\boldsymbol{f_{merged}}$ =  AdaIN($\boldsymbol{f_{src}}$,$\boldsymbol{f_{sty}^{'}}$) .
   \STATE Generate stylized source image $\boldsymbol{X_{stylized}}=Dec(0.5\boldsymbol{f_{merged}}+0.5\boldsymbol{f_{src}})$.  
   \STATE Train ASH by maximizing the loss function $\mathcal{L}_{ASH}(G,D,\boldsymbol{f_{src}},\boldsymbol{f_{sty}^{'}})$.
   \STATE Train $G$ with source domain data by minimizing segmentation loss $ \mathcal{L}_{seg}(G,\boldsymbol{X_{src}},\boldsymbol{Y_{src}})$.
   \STATE Train $G$ with stylized source domain data by minimizing adversarial loss. $\mathcal{L}_{adv}(G,D,\boldsymbol{X_{stylized}},\text{ASH})$.
   \STATE Train $D$ by minimizing adversarial loss $\mathcal{L}_{adv}(G, D,\boldsymbol{X_{stylized}},\boldsymbol{X_{src}},\text{ASH})$.
  \ENDFOR
 \end{algorithmic}
 \label{alg:workflow_alg}

\end{algorithm*}

We generate the scale $\pmb{\gamma}$ and shift $\pmb{\beta}$ coefficients from the segmentation output $G(\boldsymbol{X_{src}})$, as shown in the following equation:
\small \begin{equation}
\pmb{\gamma},\pmb{\beta}=\text{ASH}(G(\boldsymbol{X_{src}}))
\label{eqn:classwise_scaleshift}
\end{equation} \normalsize
\normalsize
We then perturb the style features $\boldsymbol{f_{sty}}$ to generate perturbed style features $\boldsymbol{f_{sty}^{'}}$:
\small \begin{equation}
\boldsymbol{f_{sty}^{'}}=\boldsymbol{Z}\cdot\boldsymbol{f_{sty}}\cdot(\pmb{\gamma}+\delta) + \pmb{\beta}
\label{eqn:perturbed style features}
\end{equation} \normalsize
where $\delta$ is a constant perturbation value \footnote{we set it as $1$}. We use a nonzero value to preserve the style features during stylization. $\boldsymbol{Z}$ is the orthogonal noise. We generate the stylized source domain images $\boldsymbol{X_{stylized}}$ with the following equation:
\small \begin{equation}
\boldsymbol{X_{stylized}} =Dec(0.5\boldsymbol{f_{src}}+0.5\text{AdaIN}(\boldsymbol{f_{src}},\boldsymbol{f_{sty}^{'}}))))
\label{eqn:stylizedsrc}
\end{equation} \normalsize
where $Dec$ is a pretrained decoder, AdaIN is the adaptive instance normalization equation defined in equation \ref{eqn:adain}.
Adversarial loss is given by the following equation:
\small \begin{equation}
\begin{aligned}
\mathcal{L}_{adv}(G,D,\text{ASH}) =& -E[\log(D(G(\boldsymbol{X_{src}})))] \\
& -E[\log(1-D(G(\boldsymbol{X_{stylized}})))]
\label{eqn:advloss}
\end{aligned}
\end{equation} \normalsize
where $G$ is the segmentation network and $D$ is the discriminator.
%\gamma_{i},\beta{i}=\text{ASH}(G(X_{s})) \\
%f_{style}^{'}=\gamma_{i}(f_{style}+1) + \beta{i} \\
%X_{stylized}=Dec(\text{AdaIN}(f_{source},f_{style}^{'})) 
We optimize the ASH module by minimizing the following loss function: 
\small \begin{equation}
\begin{aligned}
&\mathcal{L}_{ASH}(G,D,\boldsymbol{f_{src}},\boldsymbol{f_{style}^{'}}) = \\ &-\mathcal{L}_{adv}(G,D,\text{ASH})\\ &- \mathcal{L}_{c}(\boldsymbol{f_{src}},\text{AdaIN}(\boldsymbol{f_{src}},\boldsymbol{f_{sty}^{'}}))\\ &+ \mathcal{L}_{s1}(\boldsymbol{f_{sty}^{'}},\text{AdaIN}(\boldsymbol{f_{src}},\boldsymbol{f_{sty}^{'}}))\\ &- \mathcal{L}_{s2}(\boldsymbol{f_{src}},\text{AdaIN}(\boldsymbol{f_{src}},\boldsymbol{f_{sty}^{'}}))
\label{eqn:ASHloss}
  \vspace{-0.5cm}
\end{aligned}
\end{equation} \normalsize
We use the formula for content $\mathcal{L}_{c}$ and style $\mathcal{L}_{s}$ loss as defined in \cite{huang2017adain}. $\mathcal{L}_{c}$, $\mathcal{L}_{s1}$ and $\mathcal{L}_{s2}$ are described as: 
\small \begin{equation}
\mathcal{L}_{c}= \left\lVert \boldsymbol{f_{src}} - \text{AdaIN}(\boldsymbol{f_{src}},\boldsymbol{f_{sty}^{'}}) \right\rVert_{2}
\end{equation} \normalsize
\small \begin{equation}
\mathcal{L}_{s1}= \left\lVert \mu(\boldsymbol{f_{style}^{'}}) - \mu(\text{AdaIN}(\boldsymbol{f_{src}},\boldsymbol{f_{sty}^{'}})) \right\rVert_{2}
\end{equation} \normalsize

\small \begin{equation}
\mathcal{L}_{s2}= \left\lVert \mu(\boldsymbol{f_{src}^{'}}) - \mu(\text{AdaIN}(\boldsymbol{f_{src}},\boldsymbol{f_{sty}^{'}})) \right\rVert_{2}
\end{equation} \normalsize
$\mathcal{L}_{c}$ is minimized to preserve content information from the source image.
We minimize $\mathcal{L}_{s1}$ to maximize the style information retained from the style images. $\mathcal{L}_{s2}$ is maximized to minimize the style information retained from the source image. 

The segmentation network $G$ \cite{Luo_2019_CVPR} is trained to minimize segmentation loss $\mathcal{L}_{seg}$ and adversarial loss $\mathcal{L}_{adv}$. The discriminator network $D$ is trained to maximize adversarial loss $\mathcal{L}_{adv}$. Both loss functions are based on the formulation from \cite{Luo_2019_CVPR}. Segmentation loss $\mathcal{L}_{seg}$ is derived from computing the cross entropy loss for the segmentation output.

The training workflow is summarized in Algorithm \ref{alg:workflow_alg}. 
The weights for the pretrained encoder and decoder that are used during stylization are not updated during training. We only need the segmentation network for evaluation, neither the ASH module nor discriminator are required after training. 
 
\section{Experiments}
In this section, we discuss the experimental details. We first describe the datasets utilized in this work in Section. \ref{subsection:datasets}. Secondly, we provide implementation details in Section \ref{subsection:implementationdetails}. We provide details for all experimental results in Section \ref{subsection:peerstudies} - \ref{subsection:ablationstudy}. Section \ref{subsection:peerstudies} presents the performance of our approach on the benchmark datasets and compares it with the state of the art unsupervised domain adaptation and domain generalization methods. Section \ref{subsection:hyperparametereval} shows the effect of the hyperparameters on segmentation performance. Section \ref{subsection:ablationstudy} shows the ablation studies.
% shows the contribution of the different components in our approach.

\subsection{Datasets}
\label{subsection:datasets}
We use the synthetic datasets GTA5 \cite{Richter_2016_ECCV}, SYNTHIA \cite{RosCVPR16} as source domains, the real-world driving datasets Cityscapes \cite{Cordts2016Cityscapes} and Mapillary  \cite{mapillary_dataset} as the target domain. GTA5 \cite{Richter_2016_ECCV} has 24,966 images with resolution $1914 \times 1052$ pixels, while SYNTHIA \cite{RosCVPR16} has 9,400 images with $1280 \times  760$ pixels. Models are trained on the labeled source domain images and evaluated on the Cityscapes and Mapillary validation set. Similar to \cite{huang2017adain}, we use a paintings dataset (Painter by Numbers, which is derived from WikiArt) to provide 45,203 style images. 

\begin{table*}[t]
 \begin{center}
 \normalsize
 \resizebox{\linewidth}{!}{%
 \begin{tabular}{p{2cm}| c|abababababababababab|c}
  \toprule
  \multicolumn{23}{c}{\textbf{GTA5 $\rightarrow$ Cityscapes}} \\
  \toprule & \rotatebox{90}{Year} & \rotatebox{90}{Arch.}
   & \rotatebox{90}{road} & \rotatebox{90}{side.} & \rotatebox{90}{buil.} & \rotatebox{90}{wall} & \rotatebox{90}{fence} & \rotatebox{90}{pole} & \rotatebox{90}{light} & \rotatebox{90}{sign} & \rotatebox{90}{vege.} & \rotatebox{90}{terr.} & \rotatebox{90}{sky} & \rotatebox{90}{pers.} & \rotatebox{90}{rider} & \rotatebox{90}{car} & \rotatebox{90}{truck}& \rotatebox{90}{bus} & \rotatebox{90}{train} & \rotatebox{90}{motor} & \rotatebox{90}{bike} &  \rotatebox{90}{\textbf{mIoU}} \\ 
   \toprule

 Advent \cite{Vu_2019_CVPR}& 2019 &R &83.00 &1.80 &72.00 &8.20 &3.60 &16.20 &22.90 &9.80 &79.30 &17.10 &75.70 &35.10 &15.80 &70.90 &30.90 &35.30 &0.00 &16.40 &24.90 &32.60\\
 MaxSquare \cite{Chen_2019_ICCV}& 2019 &R &76.80 &14.20 &77.00 &18.80 &14.10 &14.50 &30.30 &18.00 &79.30 &11.70 &70.50 &53.00 &24.20 &68.70 &25.30 &14.00 &1.30 &20.60 &25.50 &34.60\\
 CLAN \cite{Luo_2019_CVPR}& 2019 &R &  87.20 &20.10 &77.90 &25.60 &19.70 &23.00 &30.40 &22.50 &76.80 &25.20 &76.20 &55.10 &28.10 &82.70 &30.70 &36.90 &0.80 &26.00 &17.10 &40.10  \\ 
 ASM\cite{DBLP:conf/nips/LuoLGY020}& 2020 &R & 56.20 &0.00 &7.00 &0.60 &1.00 &0.30 &0.70 &0.60 &13.80 &0.10 &0.01 &0.08 &0.04 &1.20 &0.50 &0.70 &0.20 &0.00 &0.00 & 4.40 \\
    \midrule
Domain Rand.\cite{Yue_2019_ICCV}& 2019 &R & - &- &- &- &- &- &- &- &- &- &- &- &- &- &- &- &- &- &- &\textbf{42.53}  \\
ASH (Ours)& 2021 &R & 88.30 & 19.80 & 78.80 & 23.60 & 19.50 &24.40  &30.30 &24.70 &79.10 &27.00 &74.40 &56.40 &27.90 & 83.40 &36.40 &38.40 &0.80 &22.50 &29.80 & 41.30 \\
ASH (Uni.Sem.Info.)& 2021 &R &87.40  &17.00  &77.70  &20.60  &17.80  &22.80  & 30.10 & 24.50 & 78.70 & 24.60 & 72.70 & 55.60 & 26.50 & 81.50 & 32.20 & 37.70 & 1.10 & 21.70 & 20.50& 39.50\\
  \toprule
  \multicolumn{23}{c}{\textbf{GTA5 $\rightarrow$ Mapillary}} \\
  \toprule
ASH &2021& R & 81.47 & 21.22 & 74.37 & 21.78 & 24.97 & 32.82 & 34.3 & 28.18 & 73.02 & 38.14 & 90.68 & 56.21 & 38.64 & 73.63& 37.39 & 37.27 & 7.51 & 35.9 & 36.79 & \textbf{44.44}  \\
Domain Rand.\cite{Yue_2019_ICCV} &2019& R & - & - & - & - & - & - & - & - & - & - & - & - & - & -& - & - & - & - & - & 38.05  \\
\bottomrule
\end{tabular}}
\end{center}
 \vspace{-0.5cm}
\caption{Segmentation performance of Deeplab-v2 with Resnet-101 backbone trained on GTA5, tested on Cityscapes and Mapillary. ``ASH Uni.Sem.Info"- ASH with uniform class-wise probability map (identical values across all classes and pixels).} 
\label{tab:gta-cityscapes}
\end{table*}

\begin{table*}[t]
\centering
\scriptsize

\resizebox{\linewidth}{!}{%
\begin{tabular}{p{2cm}|c|ababababababab|cc}%
\toprule
  \multicolumn{17}{c}{\textbf{SYNTHIA $\rightarrow$ Cityscapes}} \\
  \midrule &\rotatebox{90}{Year} & \rotatebox{90}{Arch.}
   & \rotatebox{90}{road} & \rotatebox{90}{side.} & \rotatebox{90}{buil.} & \rotatebox{90}{light} & \rotatebox{90}{sign} & \rotatebox{90}{vege.} & \rotatebox{90}{sky} & \rotatebox{90}{pers.} & \rotatebox{90}{rider} & \rotatebox{90}{car} & \rotatebox{90}{bus} & \rotatebox{90}{motor} & \rotatebox{90}{bike} & \rotatebox{90}{\textbf{mIoU}}& \rotatebox{90}{\textbf{mIoU16}} \\ 
   \midrule
 Advent \cite{Vu_2019_CVPR}& 2019 &R  & 72.30 &30.70 &65.20 &4.10 &5.40 &58.20 &77.20 &50.40 &10.10 &70.00 &13.20 &4.00 &27.90 &37.60& 31.80\\
 MaxSquare \cite{Chen_2019_ICCV} & 2019 &R & 57.80 &23.19 &73.63  &8.37 &11.66 &73.84 &81.92 &56.68  &20.73  &52.18  &14.71  &8.37 &39.18& 40.17 &34.96  \\
 CLAN \cite{Luo_2019_CVPR} & 2019 &R & 63.90 &25.90 &72.10 &14.30 &12.00 &72.50 &78.70 &52.70 &14.50 &62.20 &25.10 &10.40 &26.50 &40.90& 34.90 \\ 
 ASM \cite{DBLP:conf/nips/LuoLGY020} & 2020 &R  &75.40 & 18.50 &66.60 &0.10 &0.80 &67.00 &77.80 &15.60 &0.50 &11.40 &1.30 &0.03 &0.20 & 25.80 & 21.60 \\
 \midrule
Domain Rand.\cite{Yue_2019_ICCV}& 2019 &R & - &- &- &- &- &- &- &- &- &- &- &- &- &- &37.58 \\
ASH & 2021 &R & 70.20 &27.90 &75.40 &16.00&15.20 &74.20 &80.10 &55.00 &20.40 &71.10 &29.60 &10.90 &38.20 &44.90&\textbf{38.69}  \\
ASH (Segmentation Loss) & 2021 &R & 68.10 & 25.43 &74.98 &12.93 &12.98 &73.29 &78.81 &55.36 &22.13 &69.77 &30.45 &9.60 &36.75 &43.89&37.88  \\
ASH (Ground truth) & 2021 &R & 63.37 & 23.93  & 7.30&  14.58  &11.09 & 77.92& 80.60 & 54.77 & 13.42 & 68.34 & 26.49 & 12.71 & 24.73 & 42.25 & 36.29 \\

  \bottomrule
 \end{tabular} }%
 \caption{
   Segmentation performance of Deeplab-v2 with Resnet-101 backbone SYNTHIA\textrightarrow Cityscapes. ASH (Ground truth) refers to stylized images conditioned with ground truth labels. ``ASH (segmentation loss)" refers to ASH trained with segmentation loss for the stylized images $ \mathcal{L}_{seg}(G,\boldsymbol{X_{sty}},\boldsymbol{Y_{src}})$.
  } 
 \vspace{-0.5cm}
   \label{tab:synthia-cityscapes}
\end{table*}

\begin{table}
 \setlength{\tabcolsep}{12pt}
 \resizebox{\columnwidth}{!}{\begin{tabular}{lcc}
 \hline
Method & Venue &  mIoU16  \\
 \hline
 Advent \cite{Vu_2019_CVPR} & CVPR 2019  &29.33 \\
 CLAN \cite{Luo_2019_CVPR} & CVPR 2019  & 36.91 \\
Domain Rand.\cite{Yue_2019_ICCV} & ICCV 2019 & 34.12  \\
\hline
ASH & -  & \textbf{38.34} \\
ASH (segmentation loss)& - & 38.54 \\
ASH ($\delta$=0)& - & 37.61 \\
%ASH (single channel)& - & 38.87 \\
\hline
\end {tabular}}
\captionsetup{width=\columnwidth}
\caption{ Mean IoU (16 classes) for the segmentation network (Deeplab-v2 with Resnet-101 backbone) SYNTHIA \textrightarrow Mapillary.} 
\label{tab:syn-mapillary-miou}
\end{table}

\begin{figure*}[t] 
\begin{center}
\includegraphics[width=0.8\linewidth,keepaspectratio]{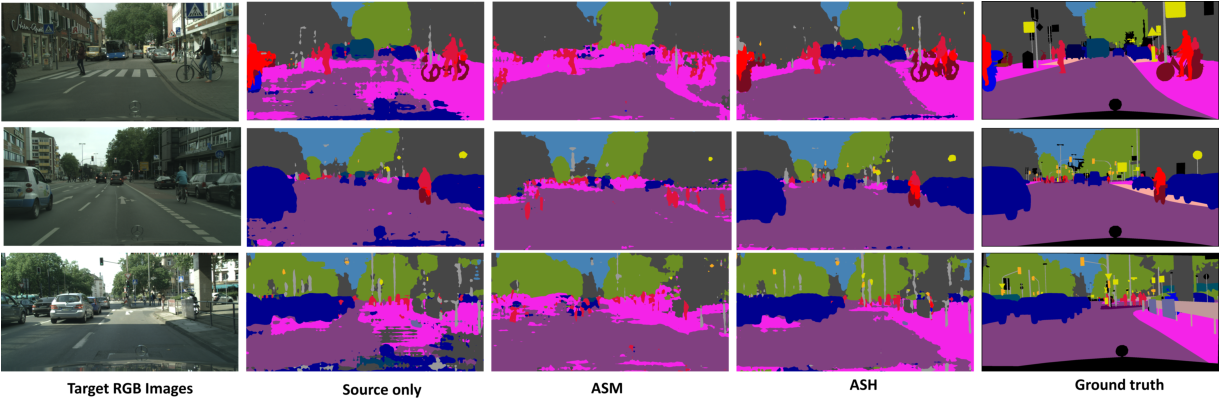}
\end{center}
 \vspace{-0.5cm}
\caption{Qualitative comparison of segmentation output for SYNTHIA \textrightarrow{} Cityscapes. For each target domain image, we show the corresponding results for ``Source only", ``ASM" Adversarial Style Mining \cite{DBLP:conf/nips/LuoLGY020}, ``ASH"(our proposed method) and the ground truth labels. }
\label{figure:qualitative }
\end{figure*}

\subsection{Implementation details}
\label{subsection:implementationdetails}
We implement our approach with the PyTorch library \cite{NEURIPS2019_9015} on a single 16GB Quadro RTX 5000. The GTA5 images are resized to $1280 \times 720$ pixels and the SYNTHIA images are resized to $1280 \times 760$ pixels. We use the Deeplab-v2 segmentation network \cite{7913730} with ResNet-101 \cite{He2015} backbone pretrained on the ImageNet dataset \cite{ILSVRC15}. The discriminator network architecture is similar to the one used in \cite{Luo_2019_CVPR}.

 We use stochastic gradient descent (SGD) to optimize the segmentation network (Deeplab-v2) and ASH module. Adam is used to optimize the discriminator network. All optimizers have a momentum of 0.9.
The initial learning rate for the segmentation network and the discriminator network is \begin{math} 2.5 \times 10^{-4} \end{math} and \begin{math} 1.0 \times 10^{-4} \end{math}. We train the network for 100,000 iterations. 
 
 \subsection{Experimental studies}
 \label{subsection:peerstudies}
 We compare our method with 5 representative methods \cite{Chen_2019_ICCV,Luo_2019_CVPR,Vu_2019_CVPR,DBLP:conf/nips/LuoLGY020,Yue_2019_ICCV} and present the results in Tables \ref{tab:gta-cityscapes} and \ref{tab:synthia-cityscapes}. \cite{Chen_2019_ICCV,Luo_2019_CVPR,Vu_2019_CVPR} are UDA approaches where target domain data are available during training; \cite{DBLP:conf/nips/LuoLGY020} aims to align domains with limited target domain data and \cite{Yue_2019_ICCV} is a domain generalization approach. Maximum Squares Loss \cite{Chen_2019_ICCV} improves upon semi-supervised learning by preventing easier classes from dominating training, CLAN \cite{Luo_2019_CVPR} seeks to reduce the difference between learned feature representations from the source and target domain, while ADVENT \cite{Vu_2019_CVPR} aims to reduce the prediction uncertainty for target domain data. ASM~\cite{DBLP:conf/nips/LuoLGY020} generates additional training data from a target domain image under one shot UDA approach. Domain randomization \cite{Yue_2019_ICCV} stylizes multiple instances of a source domain image with style images obtained from ImageNet \cite{ILSVRC15} for each iteration and performs pyramidal pooling on the extracted features to maintain feature consistency between the different stylized instances.
\begin{comment}
Previous approaches trained with a large amount \cite{Chen_2019_ICCV,Luo_2019_CVPR,Vu_2019_CVPR} or limited \cite{DBLP:conf/nips/LuoLGY020} target domain data, are all outperformed by our approach when no target domain data are available. This demonstrates the necessity of design-specific approaches. 
\end{comment}

\begin{figure*}[h]
\begin{center}
\includegraphics[width=1.0\textwidth,keepaspectratio]{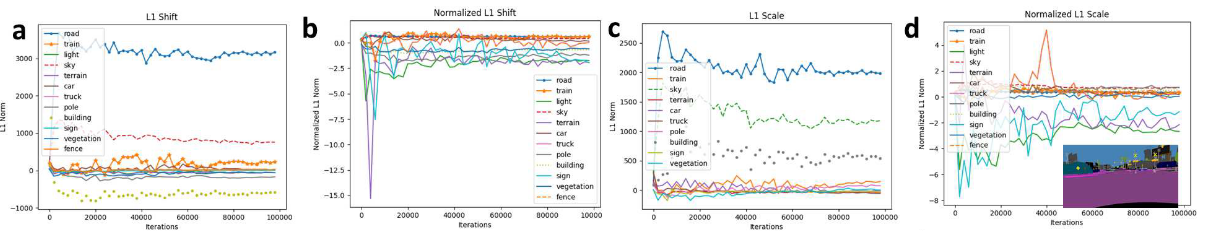}
\end{center}
 \vspace{-0.5cm}
\caption{Plots of L1 norm of shift $\boldsymbol{\beta}$ and scale $\boldsymbol{\gamma}$ coefficients versus number of iterations during training. a) Plot of L1 shift coefficients $\boldsymbol{\beta}$; b) Plot of L1 shift coefficients with class-wise normalization $\boldsymbol{\beta}$ c) Plot of L1 scale coefficients $\boldsymbol{\gamma}$; d) Plot of L1 scale coefficients $\boldsymbol{\gamma}$ with class-wise normalization. Corresponding semantic label image(inset).}
\label{figure:scale_shift_plt}
\end{figure*}

We also compare ASH with Domain Randomization (DR) \cite{Yue_2019_ICCV}, and report the results in Tables \ref{tab:gta-cityscapes}, \ref{tab:synthia-cityscapes}  and \ref{tab:syn-mapillary-miou}. ASH outperformed DR on SYNTHIA\textrightarrow Cityscapes and SYNTHIA/GTA5\textrightarrow Mapillary. There exist key differences between ASH and DR. DR generates 15 stylized images for each source domain image, while ASH only stylizes a single source domain image once per training iteration. Furthermore, DR performs spatial pyramid pooling on the extracted features. All these aspects increase computational requirements. With much less computational cost, our approach still achieves comparable results for GTA5\textrightarrow  Cityscapes and superior performance for GTA5\textrightarrow Mapillary (Table \ref{tab:gta-cityscapes}).
For a direct comparison with SFTGAN \cite{wang2018sftgan} we show results for ASH ($\delta=0$) (Table  \ref{tab:syn-mapillary-miou}). In contrast with our approach, Wang \etal \cite{wang2018sftgan} did not include a nonzero value during feature transformation in their work on super-resolution.
We observe that performance decreases when $\delta=0$. The decreased performance may be caused by loss of some style features when scale perturbation $\gamma=0$. Furthermore, Figure \ref{figure:scale_shift_plt} shows that $\gamma=0$ for some classes. The lack of stylization for these classes may have worsened performance, indicating the necessity of a nonzero $\delta$ value during stylization. \\
We also trained ASH with additional supervision (ASH segmentation loss) and show the results in Tables \ref{tab:synthia-cityscapes} and \ref{tab:syn-mapillary-miou}. We observe comparable performance with ASH. Next, we trained a ASH model that receives uniform semantic information across all classes (ASH Uni.Sem.Info) (Table \ref{tab:gta-cityscapes}). The reduced performance highlight the importance of semantic information.
Finally, we conditioned the stylization with ground truth instead of segmentation output (Table \ref{tab:synthia-cityscapes}). Segmentation performance was lowered, suggesting that the segmentation output contains useful information absent in the ground truth, which is unsurprising given the regularizing effect of soft labels during model distillation \cite{44873}.
% seen with ASH. 

\begin{table}
\setlength{\tabcolsep}{12pt}
\begin{tabular}{lllll}
\hline
 $\lambda$ &0.1 &0.01 & 0.001 &0.0001  \\
% Adversarial loss  $\lambda$ &0.1 &0.01 & 0.001 &0.0001  \\
% hyperparameter & & & &\\
\hline
mIoU   &33.33 &35.41 & 38.69 &37.53
 \\
\hline
\end{tabular}
\captionsetup{width=\columnwidth}

\caption{Segmentation performance for the segmentation task SYNTHIA\textrightarrow Cityscapes with varying adversarial loss hyperparameter magnitudes.}
\label{tab:booktabs_adv_loss_study}
\end{table}

\subsection{Hyperparameter evaluation}
\label{subsection:hyperparametereval}

In Table \ref{tab:booktabs_adv_loss_study}, we evaluate the effect of varying the adversarial loss weights on segmentation performance. Our results show that performance decreases much more when weights are increased than when the weights are decreased.

\begin{table}
\setlength{\tabcolsep}{7pt}
\centering

\begin{tabular}{c c c c c c c}
\hline
Baseline & Stylization & Orthogonal & ASH &  mIoU \\
& & Noise & & \\
\hline
\checkmark & & & & 36.6 \\
\checkmark  & \checkmark & & &  40.1  \\
\checkmark  & \checkmark & \checkmark& &  40.8 \\ 
\checkmark  & \checkmark & \checkmark&\checkmark  & 41.3\\ 
\hline
\end{tabular}
\captionsetup{width=\columnwidth}

\caption{Ablation study for GTA5\textrightarrow Cityscapes. The baseline approach is the CLAN \cite{Luo_2019_CVPR} method trained on source domain data. Stylization refers to the model trained with additional stylized data, ASH is our proposed method.}
\label{tab:booktabs_ablation}
\end{table}

\begin{table}
\centering
\setlength{\tabcolsep}{12pt}

\begin{tabular}{c c c c }
\hline
$\mathcal{L}_{c}$  & $\mathcal{L}_{s1}$  & $\mathcal{L}_{s2}$ &  mIoU \\

\hline
 & &   &36.91 \\
\checkmark & &   &37.13 \\
\checkmark  & \checkmark &  &  38.27 \\
\checkmark  & \checkmark & \checkmark&   38.69 \\ 
 
\hline
\end{tabular}
\captionsetup{width=\columnwidth}

\caption{Ablation study for the ASH sublosses from equation \ref{eqn:ASHloss}, SYNTHIA\textrightarrow Cityscapes}
\label{tab:booktabs_ablation_sublosses}
\end{table}

\subsection{Ablation study}
\label{subsection:ablationstudy}

In Table \ref{tab:booktabs_ablation}, we compare our methods with CLAN as the baseline method. 
Training with stylized images improves segmentation performance on target domain data. Since stylization varies texture information, there is less overfitting to these domain variant features. This improves the generalizability of the trained model.
Adding orthogonal noise to the style features improves performance, which could be caused by the increased diversity of the style features.
In Table \ref{tab:booktabs_ablation_sublosses}, we evaluate the effect of the different adversarial sublosses. While omitting any of the sublosses worsens performance, $\mathcal{L}_{s1}$ appears to have the greatest effect. Since $\mathcal{L}_{s1}$ determines the amount of style information retained from $X_{sty}$, this suggests that the degree of stylization greatly influences generalization performance.
% improves the segmentation performance.

\section{Discussion} 
\label{section:discussion}
\subsection{Scale and shift coefficients}
We further investigate the scale and shift coefficients by class (Figure \ref{figure:scale_shift_plt}). We calculate the L1 norm of all the scale and shift coefficients for each class in a single source image. This was obtained from the change in the L1 norm after zeroing the contribution of that class.\\
As expected, majority classes contribute more to the scale and shift coefficients than minority classes. In particular, ``road" and ``sky" classes have a larger effect on scale and shift coefficients compared to other classes such as ``pole" and ``light" (Figure \ref{figure:scale_shift_plt} a,c). Since larger scale and shift coefficients are proportionate to the change in the style features, this suggests that ``road" and ``sky" classes undergo a larger degree of stylization compared to ``pole" and ``light" classes.\\ 
These observations lead us to suggest that the ASH module selectively stylizes classes that occupy a large proportion of pixels in the predicted segmentation output for a given image (e.g ``road", ``sky"). Since the ASH module is optimized to generate stylized images that maximize adversarial loss, it appears that the network stylizes majority class pixels more than minority class pixels to maximize adversarial loss by increasing task difficulty. \\
Furthermore, several classes have negligible scale and shift coefficients. Although these classes (e.g ``vegetation", ``pole") are present in the segmentation output, regions corresponding to these classes do not undergo significant stylization compared to the majority classes. Classes such as ``vegetation" and ``pole" do not vary considerably in terms of colour information or texture. Consequently, stylizing these classes does not significantly affect the adversarial loss, which might explain the small variations in scale and shift coefficients.
\begin{comment}
Interestingly, we notice that some classes, such as ``vegetation", have shift and scale coefficients that oscillate about 0 (Figure \ref{figure:scale_shift_plt} c). This indicates that stylization in these regions will be negligible. This suggests that segmentation network performance for such classes may be unaffected by style variations.
\end{comment}
 \subsection{Normalized scale and shift coefficients}
We normalize the class-wise change in scale and shift coefficients by the number of pixels predicted for each class. This was done to provide greater clarity on the stylization for minority classes, since minority classes (e.g. ``pole", ``light") have much fewer pixels compared to majority classes (e.g. ``road", ``building").

While it may not be apparent from the plots in Figure \ref{figure:scale_shift_plt} a and c, classes that occupy a smaller area in the image also undergo stylization.
We observe that ``road" and ``building" classes have smaller absolute normalized shift and scale coefficients, while ``terrain" , ``light" and ``sign" classes have much larger absolute normalized L1 coefficients (Figure \ref{figure:scale_shift_plt} b,d). 
These results show that almost all classes, with the exception of classes such as 'vegetation', do undergo stylization, though majority classes are generally stylized to a greater extent compared to miniority classes.
\begin{comment}
Finally, We observe that the negative scale and shift coefficients for classes such as ``terrain", ``vegetation", ``light" and ``sign" become more apparent after normalization.
\end{comment}
\section{Conclusions}
In this paper, we introduce the adversarial style hallucination network, which addresses the problem of adapting to an unseen target domain. By using an adversarial approach conditioned on semantic information, ASH can adaptively stylize the source domain images. Additionally, using semantic information allows ASH to account for class-wise differences during stylization instead of treating all classes equally.  Experimental results demonstrate the efficacy of our proposed method, showing it to be competitive with state-of-the-art work.

\section{Acknowledgements}
The computational work for this article was partially performed on resources of the National Supercomputing Centre, Singapore (https://www.nscc.sg).

{\small
\bibliographystyle{ieee_fullname}
\bibliography{arxiv}
}

\end{document}